\documentclass[conference]{IEEEtran}

\usepackage{cite}
\usepackage{graphicx}

\usepackage[cmex10]{amsmath}
\usepackage{url}
\usepackage{amssymb}
\usepackage{mathtools}
\usepackage{bm}
\usepackage{algorithm,algpseudocode}
\usepackage{booktabs}
\usepackage{multirow}
\usepackage{booktabs}
\usepackage{caption}
\usepackage{subcaption}

\usepackage{comment}

\def\eg{\textit{e.g.}}

\DeclareMathOperator*{\argmin}{arg\,min}
\DeclareMathOperator*{\argmax}{arg\,max}

\def\BibTeX{{\rmfamily B\kern-.05em{\scshape i\kern-.025em b}\kern-.08em
 T\kern-.1667em\lower.7ex\hbox{E}\kern-.125em X}}

\begin{document}
\title{
Hybrid-FL for Wireless Networks: Cooperative Learning Mechanism Using Non-IID Data
}

\author{
\IEEEauthorblockN{
Naoya Yoshida\IEEEauthorrefmark{1},
Takayuki Nishio\IEEEauthorrefmark{1},
Masahiro Morikura\IEEEauthorrefmark{1},
Koji Yamamoto\IEEEauthorrefmark{1}, and
Ryo Yonetani\IEEEauthorrefmark{2}
}

\IEEEauthorblockA{\IEEEauthorrefmark{1}
Graduate School of Informatics, Kyoto University, Yoshida-honmachi, Sakyo-ku, Kyoto 606-8501, Japan
}
\IEEEauthorblockA{\IEEEauthorrefmark{1}
nishio@i.kyoto-u.ac.jp
}

\IEEEauthorblockA{\IEEEauthorrefmark{2}
OMRON SINIC X Corporation, Japan
}
\IEEEauthorblockA{\IEEEauthorrefmark{2}
ryo.yonetani@sinicx.com
}
}

\maketitle
\begin{abstract}
  This paper proposes a cooperative mechanism for mitigating the performance degradation due to non-independent-and-identically-distributed (non-IID) data in collaborative machine learning (ML), namely federated learning (FL), which trains an ML model using the rich data and computational resources of mobile clients without gathering their data to central systems.
  The data of mobile clients is typically non-IID owing to diversity among mobile clients' interests and usage, and FL with non-IID data could degrade the model performance.
  Therefore, to mitigate the degradation induced by non-IID data, we assume that a limited number (\eg, less than 1\%) of clients allow their data to be uploaded to a server, and we propose a hybrid learning mechanism referred to as Hybrid-FL, wherein the server updates the model using the data gathered from the clients and aggregates the model with the models trained by clients.
  The Hybrid-FL solves both client- and data-selection problems via heuristic algorithms, which try to select the optimal sets of clients who train models with their own data, clients who upload their data to the server, and data uploaded to the server. The algorithms increase the number of clients participating in FL and make more data gather in the server IID, thereby improving the prediction accuracy of the aggregated model.
  Evaluations, which consist of network simulations and ML experiments, demonstrate that the proposed scheme achieves a 13.5\% higher classification accuracy than those of the previously proposed schemes for the non-IID case.
\end{abstract}
\IEEEpeerreviewmaketitle

\section{Introduction}\label{sec:Intro}
The collaborative machine learning (ML), which leverages data and computation resources of mobile devices for ML model training, has attracted considerable attention. Mobile devices such as smartphones have rich data to empower modern artificial intelligence (AI) products by cutting-edge ML techniques. However, to train a modern ML model requires huge computation, and some forms of data such as e-mails and healthcare data are privacy sensitive or/and confidential. Federated learning (FL), which is a framework of the collaborative ML, solves these issues \cite{DBLP:conf/aistats/McMahanMRHA17}. The procedure of FL is similar to the existing cooperative network computing technologies such as fog and mobile cloud computing. In FL, a coordinator node, which is an edge or a cloud server usually, assigns ML tasks to mobile clients and gathers the models updated by the mobile clients using their own data, instead of gathering data and training a model by the coordinator itself. Because the clients train models, the computation load in the coordinator is small. In addition, because the data is kept in the local storage of mobile clients, confidential data can be used for training the model.

\begin{figure}[t]
  \centering
    \includegraphics[width=0.9\columnwidth]{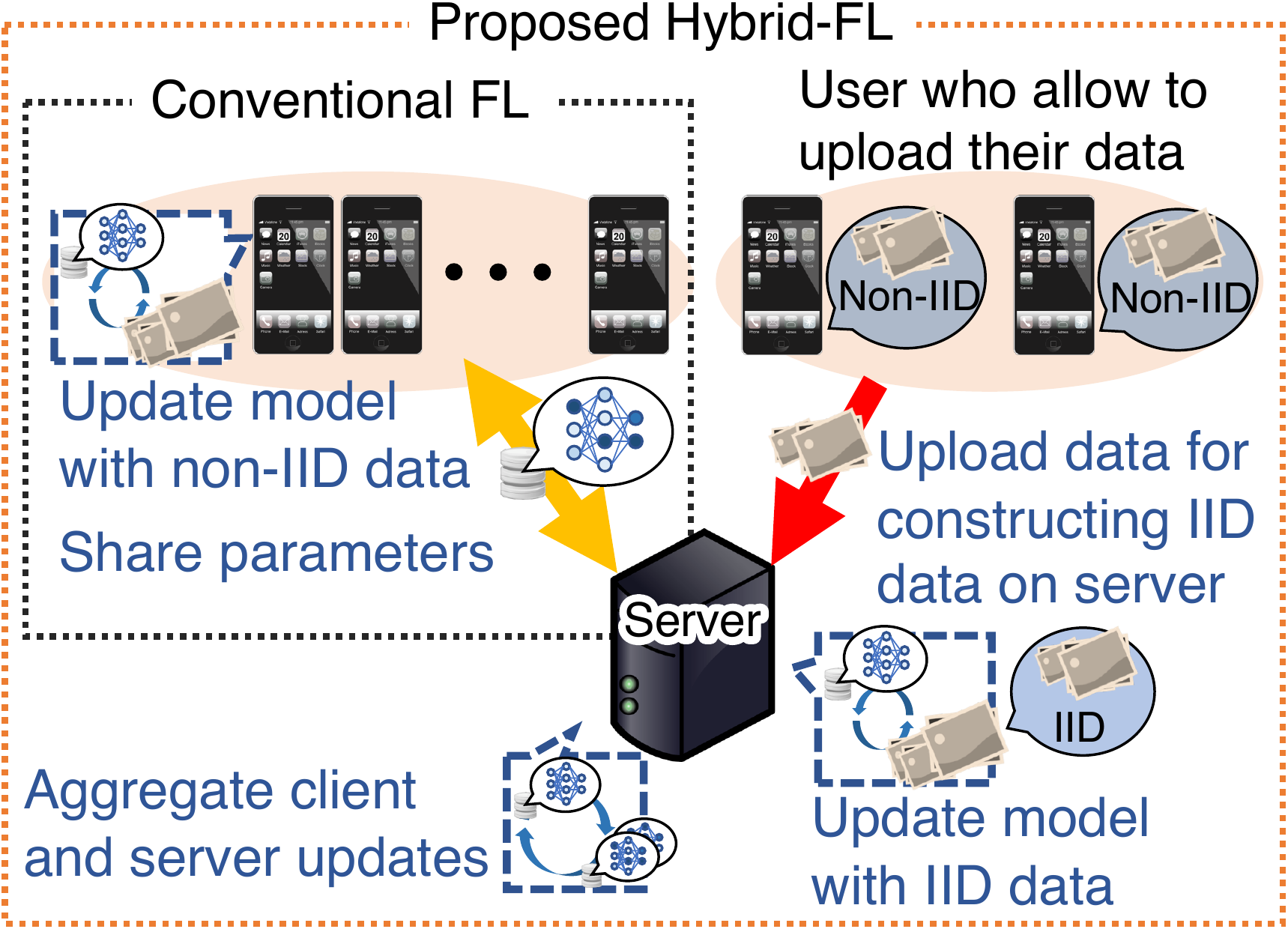}
    \caption{Proposed Hybrid-FL. In addition to the clients that update the model locally, some clients upload their own data, and the server updates the model using the uploaded data.}
    \label{fig:Hybrid-FL}
\end{figure}

A major issue in FL operations in wireless networks is the resource-scheduling problem arising from client heterogeneity and bandwidth limitation, as with fog and mobile cloud computing.
It is necessary for mobile clients with different wireless-link qualities and computation capabilities to update and communicate a few gigabyte model.
Thus, the FL operator has to coordinate regarding which client should or should not participate in FL and how much bandwidth should be allocated to each client.
Resource scheduling in fog and mobile cloud computing has been studied extensively \cite{Nishio13, Zeng16}, but resource scheduling in FL faces new challenges.  %In addition, the number of training epochs and the amount of clients' data for their updates should be coordinated in consideration with their computation power and effect for the ML model.
In FL, we have to pay attention to the distribution of client data in addition to considering the communication and computation capabilities and the amount of the client data. Because FL relies on stochastic gradient descent (SGD), which is widely employed for training deep neural networks, the population distribution must be represented by the sample training data, which is referred to as independent and identically distributed (IID) data, to provide an unbiased estimate of the full gradient \cite{10.1007/978-3-7908-2604-3_16}. However, the data of a given client typically depends on the client's interests and usage; hence, any particular client's local dataset will not be representative of the population distribution. This property is referred to as non-IID \cite{DBLP:conf/aistats/McMahanMRHA17}. In \cite{nishio2018client, zhao2018federated}, the authors reported that a non-IID data distribution could degrade the model performance of FL. The resource coordination, considering the effect for ML training, is a new problem that does not arise in the conventional cooperative network computing.

The non-IID data problem in FL has been studied in the literature \cite{zhao2018federated, sattler2019robust, wang2019adaptive}.
In \cite{zhao2018federated}, a method to share a small amount of data with other clients so that the clients' data becomes IID is proposed. However, we cannot expect that publicly available data always exists, and for security or/and data-storage reasons, clients may refuse to install unknown data on their devices.
In \cite{wang2019adaptive}, Wang et al. proposed a control algorithm that determines the best trade-off between local update and global parameter aggregation to minimize the loss function under a given resource budget and analytically derived the convergence bound for FL with non-IID data distributions.
In \cite{sattler2019robust}, a new compression framework, which works in FL with non-IID data, was proposed. However, these works did not consider resource scheduling, which is an important issue in FL in wireless networks. A few works have studied resource scheduling for the FL \cite{nishio2018client, yang2019scheduling}. These works focus on mitigating the performance degradation in ML models induced by the above-mentioned bandwidth limitation, data-amount difference, and computation and communication capability. However, the non-IID data problem in resource scheduling for FL is still an open issue.

This paper proposes a learning mechanism referred to as Hybrid-FL to mitigate the non-IID data problem. In this study, we assume a case in which a very few clients (\eg, less than 1\%) allow their data to be uploaded to an FL server. The assumption is still reasonable, particularly in the case in which the FL operators give incentives to the clients who upload their data.
In the Hybrid-FL, the server updates a model by using the data gathered from the clients in a centralized manner and subsequently aggregates the model with other models trained by distributed clients using their non-IID data.
Fig.~\ref{fig:Hybrid-FL} illustrates the Hybrid-FL concept.
By gathering data to be a good dataset, \eg, large-volume IID dataset, the performance of the aggregated model is improved compared to that of the model aggregated using only the client-updated models using non-IID data.
In addition to considering the data distribution, the traffic to gather the data is not negligible on mobile networks; therefore, we have to carefully schedule model and data uploading by considering the bandwidth limitations. In Hybrid-FL, the FL server schedules both data-uploading clients and model-uploading clients by considering the data distribution and channel condition of each client, both of which are solved using heuristic algorithms.
The Hybrid-FL protocol is based on the FL with client selection (FedCS) protocol \cite{nishio2018client}, but the scheduling algorithms and the procedures for data uploading and model training on a server are newly implemented into the FedCS protocol. The implementation requires no additional time consumption compared to that required by the FedCS protocol.
We evaluate our protocol using realistic large-scale training experiments of neural networks for image classification in a simulation environment of a cellular network. The experimental results demonstrate that the Hybrid-FL protocol achieves higher classification accuracy than that of the FedCS protocol when the clients' data is non-IID.

\begin{comment}
\subsection*{Related Work}
The concept of federated learning was first proposed in \cite{DBLP:conf/aistats/McMahanMRHA17}.
Recently, several extensions have been made to the original federated learning proposal.
For example, a method of enhancing the security of FL protocols is proposed in \cite{Bonawitz2017,Geyer2017}.
Model compression techniques for efficient communications while sacrificing model performances is proposed in \cite{Konecny2016}.
Client selection for federated learning is studied in \cite{nishio2018client}.
These studies do not consider the non-IID data distribution.
One example that considering the non-IID data distribution is \cite{zhao2018federated}, which proposed sharing a small amount of data with other clients so that the clients' data becomes IID.
However, we cannot expect that publicly available data always exists, and for security reasons, users may refuse to install unknown data on their devices.
The other one is \cite{wang2019adaptive}, which achieved the desirable trade-off between local update and global aggregation considering non-IID data distribution in order to minimize the loss function under a resource budget constraint.
However, this method do not consider heterogeneous computation and communications and/or data resources of clients.
\end{comment}

\section{System Model}\label{sec:SystemModel}
The system model follows the model in a previous work \cite{nishio2018client}.
We consider a certain MEC platform, which is located in a cellular network and consists of a server and a base station (BS).
The MEC operator manages the behaviors of the server and clients in the FL protocol.
In addition, all the training processes are assumed to be performed at midnight or in the early morning when the network is not congested, because the ML models to be trained and communicated with are typically large.
We also assumed that the MEC operator limits and manages the number of resource blocks (RBs) \cite{sesia2011lte} available for the training process.
In addition, if multiple clients communicate with the server simultaneously, then the throughput for each client decreases accordingly.

We assume that the modulation and coding schemes for radio communication is determined suitably for each client, considering its channel state and packet-loss rate to be negligible.
This leads to a different throughput for each client to communicate with the server, although the number of the allocated RBs is constant.
Even so, the channel state and throughput for each client are assumed to be stable because client devices may be unoccupied and stationary at midnight or in the early morning.

Furthermore, we consider the following additional assumptions for our proposal.
We assume that a limited number of clients allow their data to be uploaded to the server.
This assumption is reasonable because some clients will definitely agree to upload their data if incentives (\eg, a monetary reward or complimentary subscription of the application made by the FL) are provided to them.
Accordingly, if large incentives are provided, then a considerable number of clients would allow their data to be uploaded, and, consequently, we would not require to conduct FL because the server would obtain sufficient data to accurately train the model by itself. Therefore, this paper considers a case in which the ratio of the number of clients allowing to upload their data to the total number of clients is very small, \eg, less than 1\%.
Furthermore, we only consider a classification task, which is the most popular task and has wide applications.

\section{FedCS: Federated Learning with Client Selection}\label{sec:FedCS}
In this section, we briefly introduce FedCS, as presented in~\cite{nishio2018client}.
Subsequently, we describe the problem that occurs when considering a non-IID data distribution.

\begin{figure}[t]
\vspace{-3mm}
\makebox[\linewidth]{
\begin{minipage}{\linewidth}
\begin{algorithm}[H]
\floatname{algorithm}{Algorithm}
\caption{Client selection}
\label{alg:scheduling}
\begin{algorithmic}[1]
\Require{Index set of randomly selected clients $\bm{K}'$}
\State \textbf{Initialization} $\bm{S} \gets \{\}$, $t\gets 0$ %, $T_{\bm{S}=\emptyset}^\mathrm{d} \gets 0$
\While{$|\bm{K}'| > 0$}
  \State{$x \gets \argmin_{k\in \bm{K}'} f(\bm{S},k)$}
  \State{remove $x$ from $\bm{K}'$}
  \State{$t^{\prime} \gets t+T_{\mathrm{inc}}(\bm{S},x)$}
  \If{$t^{\prime}<T_{\mathrm{round}}$}
    \State{$t \gets t^{\prime}$}
    \State{add $x$ to $\bm{S}$}
  \EndIf{}
\EndWhile
\Return{$\bm{S}$}
\end{algorithmic}
\end{algorithm}
\end{minipage}
}\end{figure}

% \subsection{FedCS Protocol}
FedCS is an FL protocol that aims to work with heterogeneous clients in a practical cellular network, while mitigating the problem that occurs when some clients have limited computational resources (longer model-update times) or poor wireless channel conditions (longer model-upload times).
In FedCS, the server first randomly initializes a global model, following which the following steps are executed iteratively.
1) $\lceil K\times C \rceil$ random clients (where $K$ denotes the total number of clients, $C\in(0, 1]$ a hyper parameter representing the proportion of clients participating in each round to the total number of clients, and $\lceil\cdot\rceil$ the ceiling function) inform the MEC operator about their resource information, such as wireless-channel states, computational capacities, and data amounts relevant to the current training task (\eg, if the server is going to train a `dog-vs-cat' classifier, then the number of images containing dogs or cats).
2) Using this information, the MEC operator determines which of the clients proceed to the subsequent steps.
3) The server then distributes the parameters of the global model to the selected clients.
4) The selected clients update the global models in parallel by using their own data, and they then upload the new parameters to the server by using the RBs allocated by the MEC operator.
5) Finally, the server aggregates multiple models updated by the selected clients to improve the global model.

In Step 2), the clients are selected as shown in Algorithm~\ref{alg:scheduling}. Here,
$T_{\mathrm{inc}}(\bm{S},k)$ denotes the estimation time, which signifies for how much time a round will extend when adding the client $k$ to $\bm{S}$, and
$f(\bm{S},k)$ denotes the client-evaluation value.
In the FedCS protocol, $f(\bm{S},k)$ is $T_{\mathrm{inc}}(\bm{S},k)$.
We iteratively add the client that consumes the least time for the model upload and, subsequently, update $\bm{S}$ until the estimated elapsed time $t$ reaches the deadline $T_{\text { round }}$.
The details of estimating $T_{\mathrm{inc}}(\bm{S},k)$ are provided in~\cite{nishio2018client}.

% \subsection{Non-IID Data Problem in FedCS}
As mentioned in Section~\ref{sec:Intro}, FL including FedCS is vulnerable to the non-IID data problem.
In a practical environment, the training data of each client is typically based on the mobile-device usage of a particular client.
Therefore, the distribution of the local datasets will vary significantly among the clients.
% For example, when training a `dog-vs-cat' classifier with FL, some clients may have only dog images, while others may have only cat images.
Accordingly, in such a setting, the model performance will be significantly degraded \cite{zhao2018federated}.

\section{Hybrid Federated Learning}\label{sec:Hybrid-FL}
We propose a novel FL protocol, called Hybrid-FL, which performs efficiently for non-IID data distributions.
Hybrid-FL is a hybrid protocol derived from both centralized model training and distributed model training.
In this section, we present the proposed Hybrid-FL protocol in detail.

\begin{figure}[t]
\vspace{-3mm}
\makebox[\linewidth]{
\begin{minipage}{\linewidth}
  \begin{algorithm}[H]
    \floatname{algorithm}{Protocol}
    \caption{Hybrid-FL. Here, $K$ denotes the number of clients and $C\in(0, 1]$ the fraction of random clients that receive a resource request in each round.}
    \label{prt:Hybrid-FL}
    \begin{algorithmic}[1]
    \State \texttt{Initialization:} The server first initializes a global model either randomly or by pretraining with public data.
    \State \texttt{Resource Request:} The MEC operator then asks $\lceil K\times C \rceil$ random clients to participate in the current training task. The clients receiving the request notify the operator about their data and resource information and also whether they permit their data to be upload to the server.
    \State \texttt{Client and Data Selection:} Using this information, to complete the steps within a certain deadline, the MEC operator determines which of the clients proceed to the subsequent steps.
    It selects the sets of clients to update the models locally and then upload data to the server.
    \State \texttt{Distribution:} The server distributes the parameters of the global model to the clients selected to update the models locally.
    \State \texttt{Model Update and Data Upload:} Each set of the selected clients update the models or upload their own data for specific classes in parallel.
    \State \texttt{Scheduled Upload:} The clients selected to locally update the models upload the new parameters by using the RBs allocated by the MEC operator.
    \State \texttt{Aggregation:} The server then averages over the updated parameters, following which it replaces the global model with the averaged model.
    \State All the steps except \texttt{Initialization} are iterated multiple times until the global model achieves the desired performance or the final deadline arrives.
    \end{algorithmic}
\end{algorithm}
\end{minipage}
}\end{figure}

\begin{figure}[t]
  \centering
    \includegraphics[width=0.75\columnwidth]{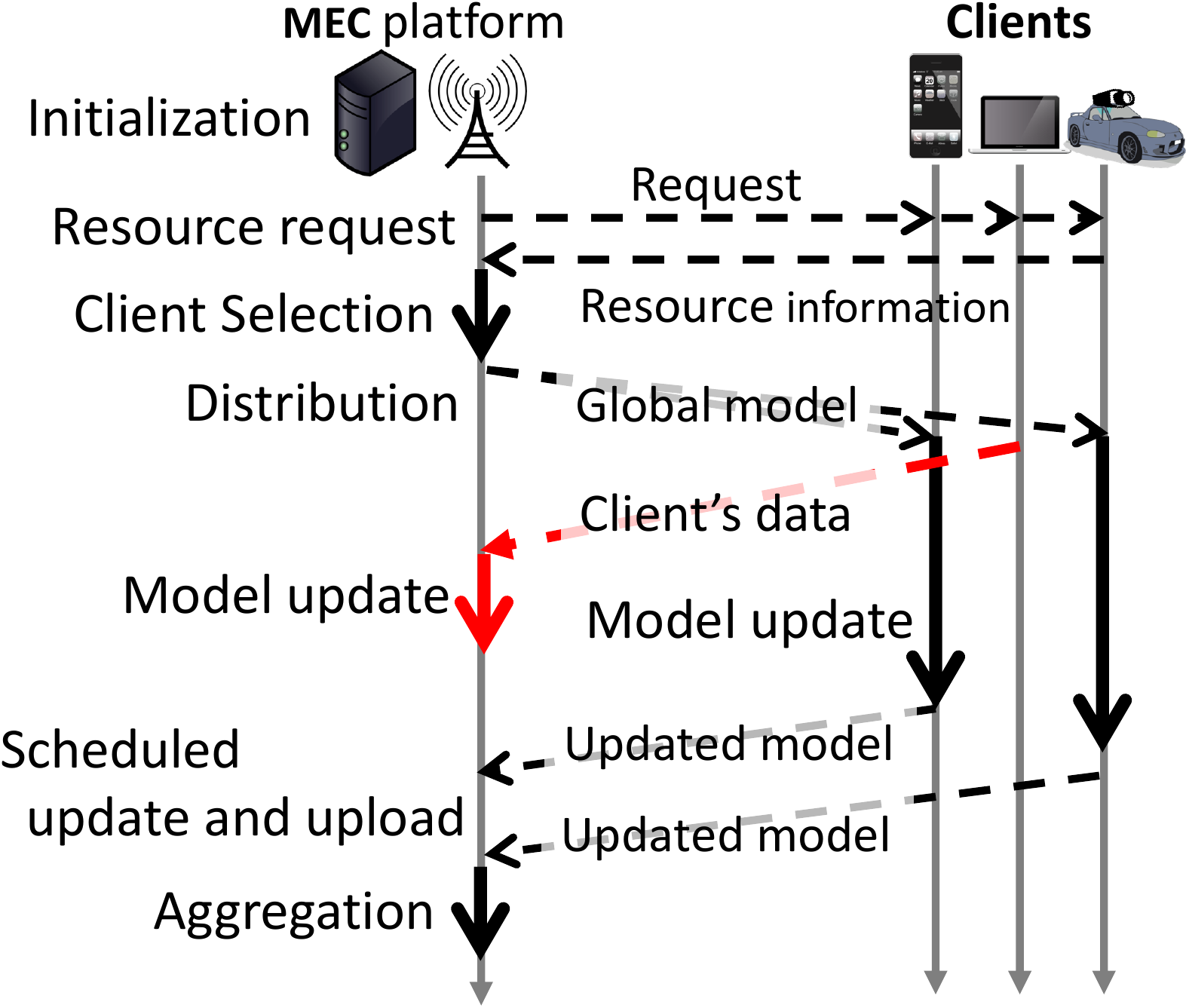}
    \caption{Overview of the Hybrid-FL protocol. The solid lines denote the computation processes, while the dashed lines indicate wireless communications. The red lines represent different points from FedCS.}
    \label{fig:diagram}
\end{figure}

\subsection{Hybrid-FL Protocol}
We present Hybrid-FL in Protocol~\ref{prt:Hybrid-FL} (see Fig.~\ref{fig:diagram} to know how each step is performed in order).
The key idea behind our protocol is that some clients upload their data to the server, and that both the server and clients update the model.
Even if each client has non-IID data, IID data can be approximately constructed on the server by combining the data stored by multiple clients.
The \texttt{Resource Request} step asks random clients to inform the MEC operator about their data amount for each class, communication resources, computational resources, and whether they permit their data to be uploaded to the server.
This information enables the operator in the subsequent \texttt{Client and Data Selection} step to estimate the time required for the \texttt{Distribution} and \texttt{Model Update and Data Upload} steps, and to determine which clients proceed to these steps.
The information is also utilized for the \texttt{Client and Data Selection} step, to select the clients that can upload data within the estimated time for each class.
In the \texttt{Client and Data Selection} step, the operator selects the sets of clients that update the models locally and those that upload data to the server.
In the \texttt{Distribution} step, a global model is distributed to the selected clients to locally update the model via multi-cast from the BS.
In the \texttt{Model Update and Data Upload} step, each set of the selected clients updates the models or uploads their own data for a specific class in parallel.
After gathering data from clients, the server updates the global model using the uploaded data.
Because the data uploads and local-model updates are performed in parallel, additional time is not required for data uploading.
Subsequently, the locally updated models are uploaded to the server in the \texttt{Scheduled Upload} step.
The server now aggregates the parameters of the updated model, and it then replaces the global model with the averaged model.
After the aggregation, the server measures the model performance using the validation data.
All the steps except \texttt{Initialization} are iterated multiple times until the final deadline arrives.

\subsection{Client and Data Selection in Hybrid-FL}
We should apply two selection methods in the Hybrid-FL protocol: one to select clients to locally update the model and the other to select the data to be uploaded.
In this section, we explain these two selections, namely client selection and data selection.

% \subsection{Selection of Clients to Locally update model}
First, we explain how the MEC operator selects clients to locally update the model.
We have two methods available to perform client selection.
The first method is the same as that explained in Section~\ref{sec:FedCS}.
The maximum possible number of clients who can complete the training process within a certain deadline are selected.
The second method selects clients such that the amounts of data of each class utilized for updating the models have close values.
This method may make Hybrid-FL more robust to non-IID data.
Furthermore, this method selects clients as described below.
Let $\bm{S}_r$ be a sequence of indices of clients selected to locally update the model in the $r$-th round, and let $\bm{N}_r=\{n_1,\cdots,n_l,\cdots,n_L\}$ be the total amount of data for each class stored by the clients indexed by $\bm{S}_1,\cdots,\bm{S}_r$.
We evaluate the bias of the data by using the coefficient of variation of $\bm{N}_r$ as follows:
\begin{align}
  \mathrm{CV}(\bm{N}_r)=\frac{\sum_{l=1}^L(n_l-\overline{n})^2/L}{\overline{n}},
\end{align}
where $\overline{n}=\sum_{l=1}^L n_l/L$.
In the $r$-th round, the \texttt{Client and Data Selection} step selects clients to locally update the model as shown in Algorithm~\ref{alg:scheduling}, where $f(\bm{S},k)=T_{\mathrm{inc}}(\bm{S},k)\cdot\mathrm{CV}(\bm{N}_r)$.
This algorithm can decrease the time required for both model uploading and updating.
It can also reduce the bias of the data used to update the models.

% \subsection{Selection of Data to Upload}
Next, we explain how the MEC operator selects the data to be uploaded within a limited time.
In the \texttt{Client and Data Selection} step, the MEC operator estimates the time required to update the model, following which it selects clients to locally update the models, as described above.
In Hybrid-FL, the clients can upload their data until the new parameters start to be uploaded.
The MEC operator selects data that can be uploaded within this time for each class from the data stored by the clients who had permitted data uploading.
We have two methods available to select the data.
The first method aims to maximize the amount of data in the server.
This method simply asks the clients having high throughputs to upload their data in order.
The second method aims to construct IID data on the server.
This method selects data as shown in Algorithm~\ref{alg:iid}.
Let $t^{\mathrm{UD}}$ denote the estimated time within which data is uploaded, $\bm{D}^\mathrm{UL}$ a set of indices of the data uploaded to the server in the \texttt{Model Update and Data Upload} step, and $t_{\bm{D}^\mathrm{UL}}$ the time required to upload the data indexed by $\bm{D}^\mathrm{UL}$.
Furthermore, let $\bm{U}$ be a set of indices describing $U$ clients that have permitted data uploading.
Note that if a client has already uploaded all the stored data in the previous rounds, then they are excluded from $\bm{U}$.
Let $\theta^{avg}_k$ , where $u\in\bm{U}$, be the throughput of client $u$, and let $\bm{D}_u$ be the data held by client $u$.
Then, $\bm{d}_{ul}\in\bm{D}_u$, where $l=1,\cdots,L$, denotes the class $l$ of the data stored by client $u$, and $L$ denotes the number of classes in the classification problem.
We iteratively add the data of each class to $\bm{D}^\mathrm{UL}$ in order until $t_{\bm{D}^\mathrm{UL}}$ reaches $t^{\mathrm{UD}}$.
Similar amounts of data will be present on the server for each class.

\begin{figure}[t]
  \vspace{-3mm}
  \makebox[\linewidth]{
  \begin{minipage}{\linewidth}
  \begin{algorithm}[H]
  \floatname{algorithm}{Algorithm}
  \caption{Selection algorithm for the data to be uploaded.}
  \label{alg:iid}
  \begin{algorithmic}[1]
      \Require $t^{\mathrm{UD}},\ \bm{U},\ \bm{D}_u$
      \State \textbf{Initialization:} $\bm{D}^\mathrm{UL} \gets \emptyset$, $flag\gets$True
      \While{$flag$}
          \For{$l$ in $1,\cdots,L$}
              \State $x\gets \argmax_{u\in\bm{U}} \theta^{avg}_u$ where $\bm{d}_{ul}\neq\emptyset$
              \State $d\gets$ the first element of $\bm{d}_{xl}$
              \If{$t_{\bm{D}^\mathrm{UL}\cup d}\leq t^{\mathrm{UD}}$}
                  \State add $d$ to $\bm{D}^\mathrm{UL}$
                  \State remove $d$ from $\bm{d}_{xl}$
              \EndIf
              \If{$t_{\bm{D}^\mathrm{UL}\cup d}> t^{\mathrm{UD}}$ or $\bm{D}_u=\emptyset,\ \forall u\in\bm{U}$}
              \State $flag\gets$False
              \EndIf
          \EndFor
      \EndWhile\\
      \Return $\bm{D}^\mathrm{UL}$
  \end{algorithmic}
  \end{algorithm}
  \end{minipage}
  }\end{figure}

\section{Performance Evaluation}\label{sec:Evaluation}
As a proof of concept that our protocol works effectively, we simulated an MEC environment and then conducted experiments by performing realistic ML tasks using publicly available large-scale datasets.
Both the simulation and experiment are inspired from \cite{nishio2018client}.
We evaluated the performance of our protocol using an IID data distribution and various non-IID data distributions.

\subsection{Simulation Settings} \label{subsec:SimulationSettings}
We simulated an MEC environment implemented in an urban microcell.
The MEC environment comprised an edge server, a BS, and $K=1000$ clients.
Furthermore, $K_{\mathrm{UL}}$ clients, which constituted $r_{\mathrm{UL}}$ (\eg, less than 0.01) of the total number of clients, permitted the uploading of their own data to the server.
The $K_{\mathrm{UL}}$ clients were randomly determined from the total of 1000.
In addition, the BS and server were co-located at the center of the 2-km radius cell, and the clients were uniformly distributed in the cell.
The computation capability of the server was sufficiently high compared to that of the clients.
Therefore, the time required both for \texttt{Client and Data Selection} and \texttt{Aggregation} could be ignored.
Similarly, the model-update time in the server could also be ignored.
We set hyperparameter $C=0.1$ and the final deadline $T_{\mathrm{final}}=400$\,min.
% We selected six clients to locally update the model in each round.

Wireless communications were modeled on the basis of long-term evolution (LTE) networks with an urban-channel model defined in the ITU-R M.2135-1 Micro NLOS model with a hexagonal-cell layout~\cite{ITU-R}.
Furthermore, we set the model parameters as follows.
The carrier frequency was 2.5\,GHz; the antenna heights of the BS and clients were 11 and 1\,m respectively; the transmission power and antenna gain of the BS and clients were 20 and 0\,dBi, respectively.
As a practical bandwidth limitation, we assumed that 10 RBs, which corresponded to a bandwidth of 1.8 MHz, were assigned to a client in each time slot of 0.5\,ms.
The throughput model was based on the Shannon capacity with a certain loss used in~\cite{6834753} with $\Delta = 1.6$ and $\rho_\mathrm{max}=4.8$.
The mean and maximum throughput of a client were 1.4 and 8.6\,Mbit/s, respectively, both of which are realistic values in an LTE network.
We considered the throughput $\theta^{avg}_k$ obtained from the above-mentioned model as the average throughput of each client, and we used this throughput in \texttt{Client and Data Selection}.
% As mentioned in Section \ref{sec:SystemModel}, all of the FL processes were assumed to be performed during the network condition was stable and client devices were likely to be unused.
% This allowed us to regard the average throughput as stable.
% Nevertheless, to consider a small variation of short-term throughput at \texttt{Data Upload} and \texttt{Scheduled Upload} that can happen in practice, everytime when clients communicate with the server we sampled the throughput from a truncated normal distribution derived from a normally distribution by limiting the random variable to one standard deviation away from the mean.
% The mean and variance are given by the average throughput and its $r_{\mathrm{var}}$\% value, respectively.
Similar to the case of \cite{nishio2018client}, to consider a fluctuation in short-term throughput, the throughput in the simulation was sampled from a truncated normal distribution from $(1-r_{\mathrm{var}}) \theta^{\mathrm{ave}}_k$ to $(1+r_{\mathrm{var}}) \theta^{\mathrm{ave}}_k$,  where $r_{\mathrm{var}}$ denotes a variance parameter.

The computation capability was modeled by how many data samples it could process in a second to update a model; the computation capability could be fluctuated because of other computation load.
Subsequently, we randomly determined the average computation capability of each client $\gamma^{\mathrm{ave}}_k$ from the range of 10 to 100, and the capability values were used for \texttt{Client and Data Selection} algorithms.
Similar to the calculation of communication capability, the computation capability in the simulation was sampled from a truncated normal distribution from $(1-r_{\mathrm{var}}) \gamma^{\mathrm{ave}}_k$ to $(1+r_{\mathrm{var}}) \gamma^{\mathrm{ave}}_k$.
% As a result, each update time used in \texttt{Client and Data Selection} varied from 5 to 500 seconds.
% In \texttt{Data Upload} and \texttt{Scheduled Upload}, the computation capability is determined by the truncated normal distribution in the same way as throughput mentioned in Section~\ref{subsec:SimulationSettings}.

\subsection{Experimental Setup for ML Tasks}
We adopted two realistic object-classification tasks using large-scale image datasets in the simulated MEC environment.
The first dataset was CIFAR-10 \cite{krizhevsky2009learning}, which is a classic object-classification dataset comprising 50,000 training images and 10,000 testing images, with 10 object classes.
This dataset has been commonly employed in many other FL studies as well \cite{DBLP:conf/aistats/McMahanMRHA17,Konecny2016}.
The second dataset was Fashion MNIST \cite{Xiao2017FashionMNISTAN}, which consists of 60,000 training images and 10,000 testing images of 10 different fashion products, such as t-shirts and bags.
This dataset also has been employed in many other FedCS studies \cite{nishio2018client}.

Furthermore, our model was a standard convolutional neural network, which was the same as that employed in \cite{nishio2018client}.
It consisted of six $3\times 3$ convolution layers (32, 32, 64, 64, 128, and 128 channels, each of which was activated using ReLU and was batch normalized, and every two of which were followed by $2\times 2$ max pooling), followed by three fully connected layers (512 and 192 units activated using ReLU and another 10 units activated using soft-max).
When updating models, we selected the following hyperparameters according to the work in \cite{nishio2018client}: $50$ for mini-batch size, $5$ for the number of epochs in each round, $0.25$ for the initial learning rate of SGD updates, and 0.99 for learning-rate decay.

\subsection{Data Distribution}
The training data, which comprised images of 10 classes, were distributed among $K=1000$ clients.
Let $r_l$ denote the ratio of clients with data on $l$ classes to the number of all the clients, and $\bm{r}=\left(r_1,r_2,\dots,r_{10}\right)$. $K \cdot r_l$ clients had images for $l$ classes. Because the images are from 10 different classes, we have $\sum_{l=1}^{10} r_l=1$.
Furthermore, the number of images owned by each client was randomly determined in a range from 100 to 1,000. Each client then sampled the specified number of images randomly from different subsets, where $l$ of the 10 classes were randomly selected.
For example, when $r_{10}=1$ and $r_1,\dots ,r_{9}=0$, the clients could have data from all the classes, thereby representing an IID data distribution.
However, when $r_1=1$ and $r_2,\dots ,r_{10}=0$, the clients could have data from only one class.
We set $\bm{r}$ to follow a truncated normal distribution.
Let $\mathbb{R}$ be the set of real numbers; $\mu, \sigma, a, b\in\mathbb{R}$; and $a\leq\mu\leq b$.
A cumulative-distribution function of the truncated normal distribution for $a\leq x\leq b$ is given by
\begin{align}
    F(x ; \mu, \sigma, a, b)=\frac
    {\Phi\Bigl(\frac{x-\mu}{\sigma}\Bigr)-\Phi\Bigl(\frac{a-\mu}{\sigma}\Bigr)}
    {\Phi\Bigl(\frac{b-\mu}{\sigma}\Bigr)-\Phi\Bigl(\frac{a-\mu}{\sigma}\Bigr)}.
\end{align}
Furthermore, $\Phi(\cdot)$, which denotes the cumulative-distribution function of the standard normal distribution, is given by
\begin{align}
  \Phi(x)=\frac{1}{2}\left(1+\operatorname{erf}\!\left(x / \sqrt{2}\right)\right).
\end{align}
We fixed $a=0.5,\,b=10.5$, and set $r_l\,(l=1,\dots,10)$ for various $\mu$ and $\sigma$ as
\begin{align}
  r_l=F(l+0.5)-F(l-0.5).
\end{align}
Thus, the smaller the value of $\mu$ is, the greater is the number of clients that have data from limited classes, the larger is the value of $\sigma$, and the greater is the variety of clients, with some of the clients having data from various classes while others having data from limited classes.

% \begin{figure}[t]
%   \centering
%   \subcaptionbox{CIFAR-10.}{
%       \includegraphics[width=0.8\columnwidth]{./fig/cifar07_resourcefix1.pdf}
%   }
%   \caption{Accuracy achieved by FedCS and Hybrid-FL trained on CIFAR-10 when throughput and computational resources are not stable.}
%   \label{fig:mugraph-resouceFlexible}
% \end{figure}

% \begin{figure}[t]
%   \centering
%   \subcaptionbox{CIFAR-10.}{
%       % \includegraphics[width=0.8\columnwidth]{./fig/cifar07.pdf}
%       \includegraphics[width=0.78\columnwidth]{./fig/cifar07_OnlyServer.pdf}
%   }
%   \subcaptionbox{Fashion MNIST.}{
%       % \includegraphics[width=0.8\columnwidth]{./fig/fashion07.pdf}
%       \includegraphics[width=0.78\columnwidth]{./fig/fashion07_OnlyServer.pdf}
%   }
%   \caption{Effect of data non-IIDness on the prediction accuracy of models trained using FedCS and Hybrid-FL for CIFAR-10 and Fashion MNIST ($\sigma=0.7$). A smaller $\mu$ value means that clients tend to have data from a lower number of classes.}
%   \label{fig:mugraph}
% \end{figure}

\begin{figure}[t]
  \centering
  \includegraphics[width=1\columnwidth]{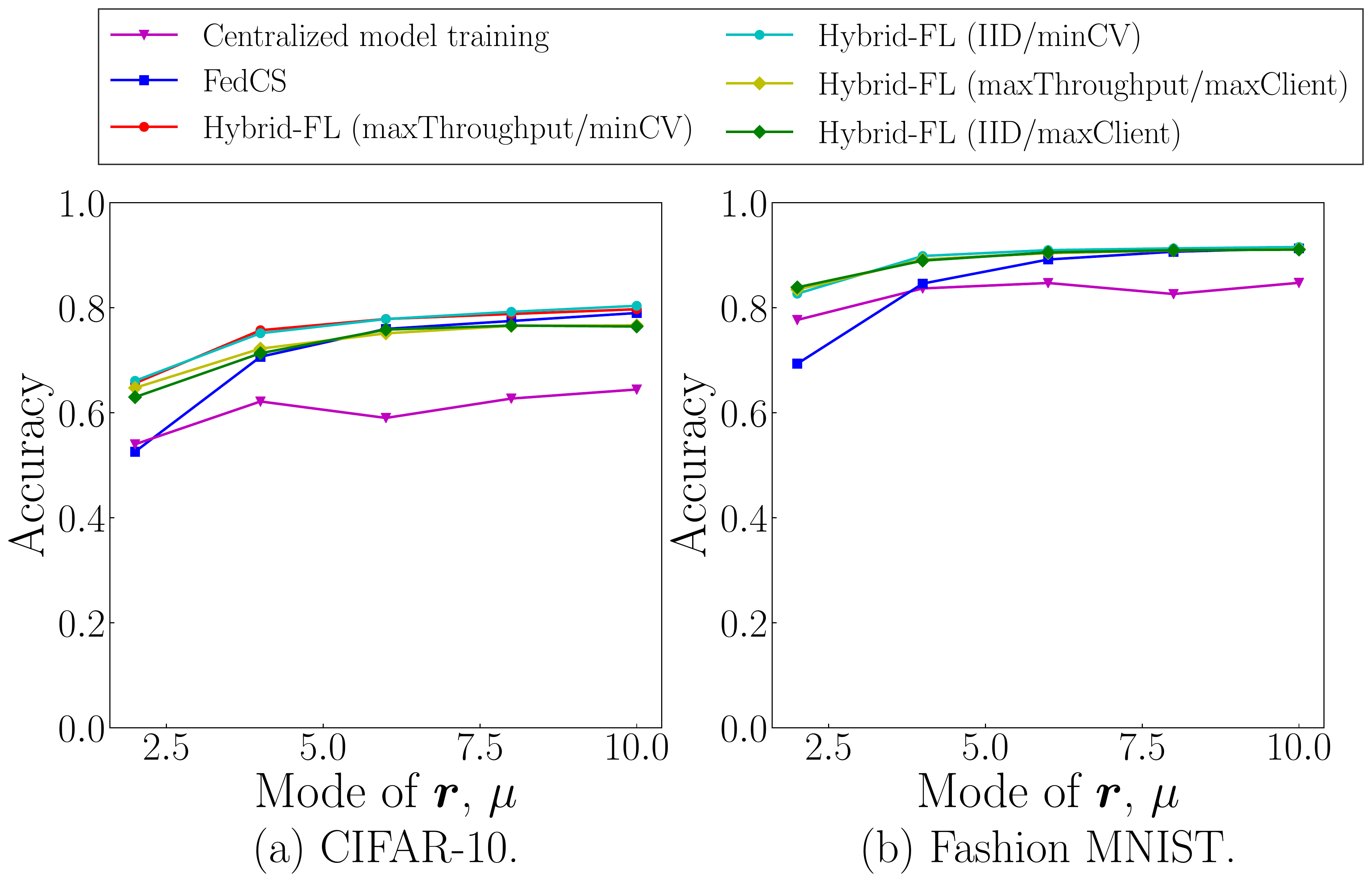}
  \caption{Effect of data non-IIDness on the prediction accuracy of models trained using FedCS and Hybrid-FL for CIFAR-10 and Fashion MNIST ($\sigma=0.7$). A smaller $\mu$ value means that clients tend to have data from a lower number of classes.}
  \label{fig:mugraph}
\end{figure}

\subsection{Evaluation Results}
We evaluated four variations of the Hybrid-FL protocol, representing the combinations of two methods each for selecting data-uploading clients and model-uploading clients: The selection of data-uploading clients was based on the throughput, referred to as \emph{maxThroughput}, or based on Algorithm~\ref{alg:iid}, referred to as \emph{IID}; the selection of model-uploading clients was based on $T_{\mathrm{inc}}(\bm{S},k)$, referred to as \emph{maxClient}, or based on $T_{\mathrm{inc}}(\bm{S},x) \mathrm{CV}(\bm{N}_r)$, referred to as \emph{minCV}.

We compared the Hybrid-FL protocol with the FedCS protocol, which does not utilize data uploading.
We also evaluate centralized model training, which was a result when the server trained a model by using the data obtained from model-uploading clients.
Subsequently, we evaluated the mean accuracy for the last 100\,minutes (from $T=300$ to $T=400$\,minutes), where the mean accuracy of each method was averaged over 10 trials.

\begin{table}[t]
  \vspace{1mm}
  \caption{Effect of the ratio of clients with different data non-IIDness on prediction accuracy of the models trained using FedCS and Hybrid-FL for CIFAR-10 and Fashion MNIST ($\mu=4$).}
  \label{table:sigtable}
  \centering
  \begin{tabular}{lccr}
      \toprule
      \multirow{2}{*}{Method} & \multicolumn{3}{c}{CIFAR-10} \\
      \cline{2-4}
        & $\sigma=0$    & $\sigma=0.7$   & $\sigma=\infty$   \\
      \midrule
      FedCS             &0.716  & 0.707 & 0.733  \\
      \midrule
      \textbf{Hybrid-FL} &   &  &   \\
      \, maxThroughput/minCV     & \textbf{0.765} & \textbf{0.756} & 0.765  \\
      \, IID/minCV         & 0.762 & 0.750 & \textbf{0.771}  \\
      \, maxThroughput/maxClient    & 0.727 & 0.731 & 0.735  \\
      \, IID/maxClient     & 0.726  & 0.721 & 0.745 \\
      \midrule
      \midrule
      \multirow{2}{*}{Method} & \multicolumn{3}{c}{Fashion mnist} \\
      \cline{2-4}
        & $\sigma=0$    & $\sigma=0.7$   & $\sigma=\infty$   \\
      \midrule
      FedCS             & 0.850  & 0.846 & 0.891   \\
      \midrule
      \textbf{Hybrid-FL} &   &  &   \\
      \, maxThroughput/minCV     & 0.894 & 0.894 & 0.901  \\
      \, IID/minCV         & \textbf{0.898} & \textbf{0.899} & \textbf{0.906}  \\
      \, maxThroughput/maxClient    & 0.894 & 0.892 & 0.897  \\
      \, IID/maxClient     & 0.894  & 0.890 & 0.900 \\
      \bottomrule
  \end{tabular}
  \label{table:result}
  \end{table}

\textbf{Effect of Non-IID Data: }
Fig~\ref{fig:mugraph} illustrates the accuracy as a function of $\mu$ when $\sigma=0.7$, $r_{\mathrm{UL}}=0.01$ and $r_{\mathrm{var}}=0$.
For small $\mu$, where many clients have data from a few classes, the prediction accuracies of all the methods decreased, but Hybrid-FL maintained its accuracy higher than that of FedCS for both the CIFAR-10 and Fashion MNIST tasks.
Specifically, Hybrid-FL (IID/minCV) with $\mu=2$ achieved a 13.5\% and 12.5\% higher accuracy for the CIFAR-10 and Fashion MNIST tasks, respectively.
Centralized model training achieved the worst accuracy for the CIFAR-10 task because the server cannot obtain sufficient data to train accurate model.
Comparing the variations of Hybrid-FL,
Hybrid-FL (IID/minCV) and Hybrid-FL (maxThroughput/minCV) yielded similar performances and outperformed the other variations.
We expect several reasons responsible for these results. First, as expected, the strategy for selecting model-uploading clients reduced the imbalance of models aggregated to the global model. Second, the strategy for selecting data-uploading clients might not make a significant impact because the throughput-based selection became random sampling owing to the randomly determined throughput, thereby generating approximately IID data on the server. In addition, the maxClient selection tended to select the clients that required a very short time to update the model, and, thus, the available time for uploading the data became shorter than that in the minCV selection. We confirmed the amount of data gathered on the server for the minCV selection to be approximately 10\% larger than that in the maxClient selection. This increase in the amount of approximately IID data improved the model performance.

Subsequently, the effect of the coexistence of clients who have data with different non-IIDness which was determined using $\sigma$ of the data distribution.
Table~\ref{table:result} presents the accuracy for different $\sigma$ with $\mu=4$ on the CIFAR-10 and Fashion MNIST tasks, where $\sigma=\infty$ and $\sigma=0$ imply that $r_1 = ... = r_{10} = 0.1$ and $r_{\mu}=1$ with the others being zero, respectively.
These results demonstrate that Hybrid-FL achieved higher accuracy than that of the conventional method irrespective of the coexistence of clients who had data from different number of classes.

\begin{figure}[t]
  \centering
      \includegraphics[width=0.75\columnwidth]{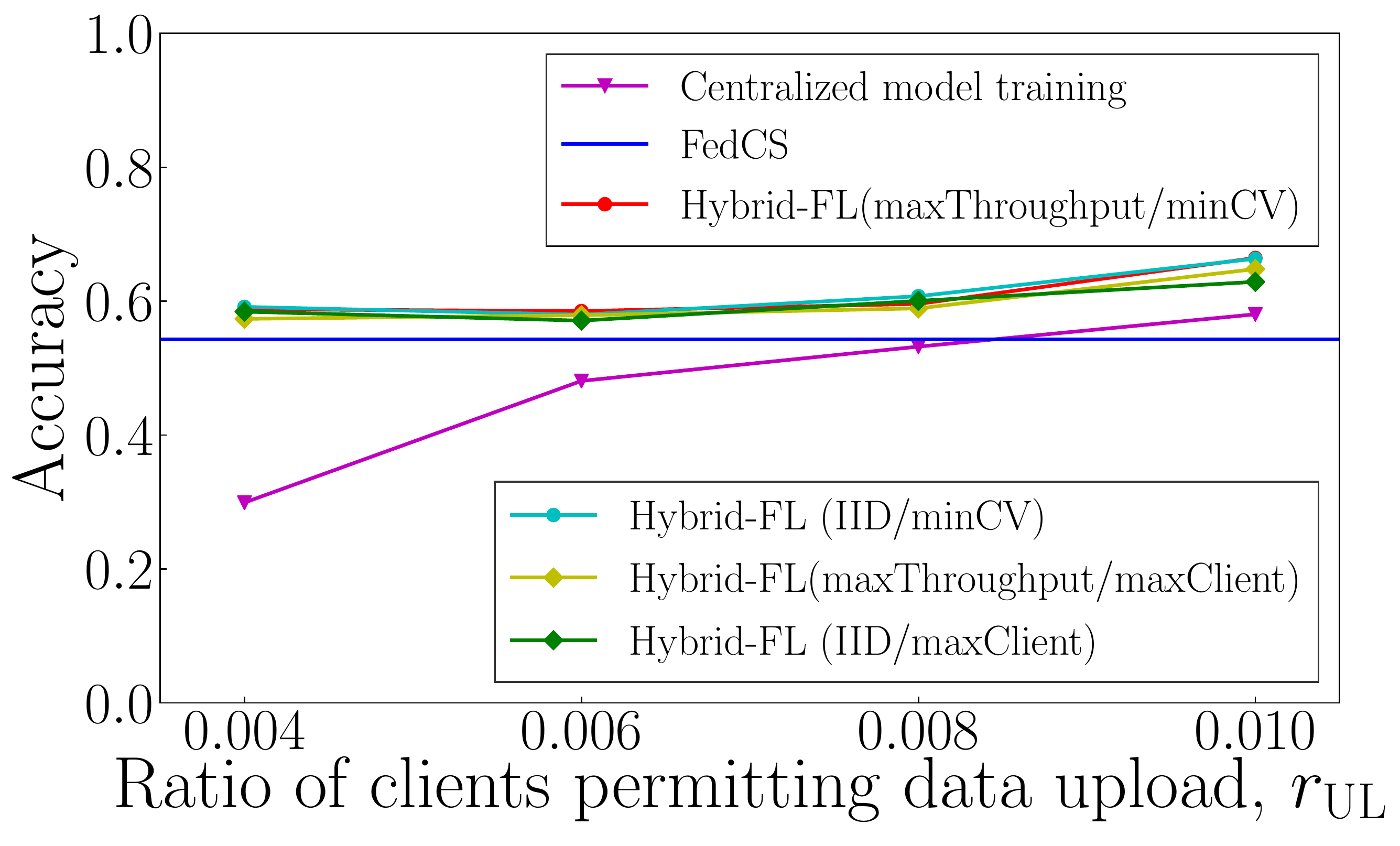}
  \caption{Effect of the number of data-uploading clients on prediction accuracy of models trained using FedCS and Hybrid-FL for CIFAR-10 task ($\sigma=0.7, \mu=2$).}
  \label{fig:Upratio}
\end{figure}

\textbf{Effect of the number of data-uploading clients: }
Fig~\ref{fig:Upratio} illustrates the accuracy as a function of $r_{\mathrm{UL}}$ on the CIFAR-10 task.
We set the data-distribution parameters as $\sigma=0.7, \mu=2$, and $r_{\mathrm{var}}=0$.
Upon increasing $r_{\mathrm{UL}}$, the prediction accuracy of both the Hybrid-FL and centralized model training increased.
Furthermore, when $r_{\mathrm{UL}}$ was less than $0.01$, the Hybrid-FL achieved higher accuracy than those of the existing methods.
We expect that centralized model training will probably achieve the highest accuracy because the server can obtain sufficient data to train accurate model by itself when $r_{\mathrm{UL}}$ is sufficiently large.

\begin{figure}[t]
  \centering
      \includegraphics[width=0.75\columnwidth]{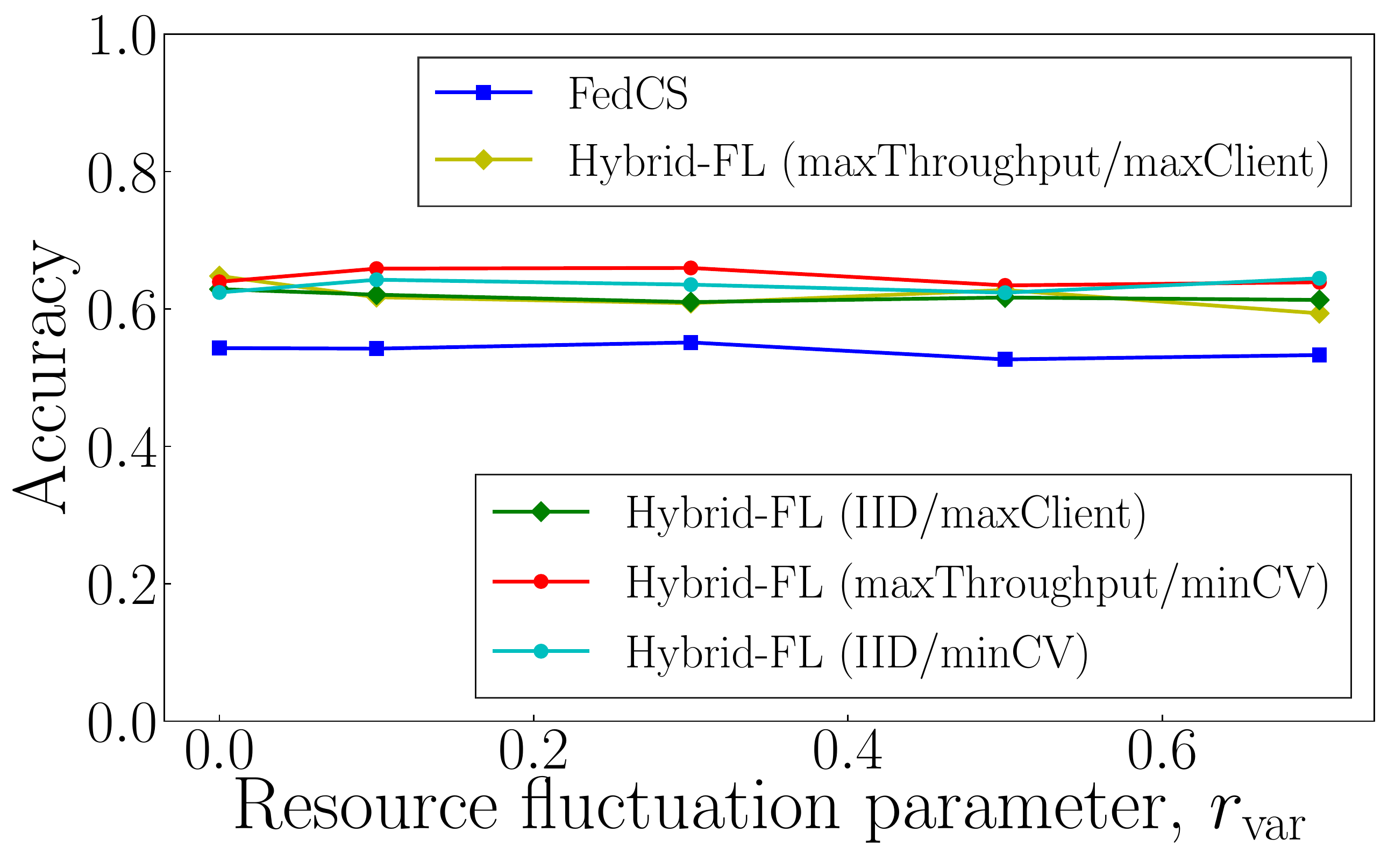}
  \caption{Accuracy achieved by Hybrid-FL trained on CIFAR-10 ($\sigma=0.7, \mu=2$) with different resource-fluctuation parameters.}
  \label{fig:ResourceFluctuation}
\end{figure}

\textbf{Effect of the resource-fluctuation parameter $r_{\mathrm{var}}$ : }
In the above-mentioned evaluations, we set $r_{\mathrm{var}}$ to $0$, meaning that the resources are stable and do not change during an FL round. Here, we show the results when the resources can be changed from a value used for the selection algorithms.
Fig~\ref{fig:ResourceFluctuation} illustrates the model accuracy for different resource-fluctuation parameters, $r_{\mathrm{var}}$, for CIFAR-10 when $\sigma=0.7, \mu=2$, and $r_{\mathrm{UL}}=0.01$.
For large $r_{\mathrm{var}}$, the prediction accuracies for all the methods decreased because the selection results were getting away from selections, maximizing their utility functions because of the gap among the resource values. However, these methods can still train ML models.

% \begin{table}[t]
% \caption{Accuracy achieved by FedCS and Hybrid-FL trained on CIFAR-10 and Fashion MNIST ($\mu=4$). The parentheses of Hybrid-FL protocol indicates selection algorithm of clients to upload data and selection algorithm of clients to locally update models respectively.}
% \label{table:sigtable}
% \centering
% \begin{tabular}{lcccr}
%     \toprule
%     \multirow{2}{*}{Method} & \multicolumn{3}{c}{CIFAR-10} \\
%     \cline{2-4}
%       & $\sigma=0$    & $\sigma=0.7$  & $\sigma=2$ & $\sigma=\infty$   \\
%     \midrule
%     FedCS             &0.716  & 0.707 &0.707 & 0.733  \\
%     \midrule
%     \textbf{Hybrid-FL} &   &  &   \\
%     \, Amount/IID     & \textbf{0.765} & \textbf{0.756} &\textbf{0.764}& 0.765  \\
%     \, IID/IID         & 0.762 & 0.750 &0.760& \textbf{0.771}  \\
%     \, Amount/Amount    & 0.727 & 0.731 &0.714& 0.735  \\
%     \, IID/Amount     & 0.726  & 0.721 &0.723& 0.745 \\
%     \midrule
%     \midrule
%     \multirow{2}{*}{Method} & \multicolumn{3}{c}{Fashion mnist} \\
%     \cline{2-4}
%       & $\sigma=0$    & $\sigma=0.7$   && $\sigma=\infty$   \\
%     \midrule
%     FedCS             & 0.850  & 0.846 &0.856& 0.891   \\
%     \midrule
%     \textbf{Hybrid-FL} &   &  &   \\
%     \, Amount/IID     & 0.894 & 0.894 &0.897& 0.901  \\
%     \, IID/IID         & \textbf{0.898} & \textbf{0.899} &0.899& \textbf{0.906}  \\
%     \, Amount/Amount    & 0.894 & 0.892 &0.714& 0.897  \\
%     \, IID/Amount     & 0.894  & 0.890 &0.889& 0.900 \\
%     \bottomrule
% \end{tabular}
% \label{table:result}
% \end{table}

\section{Conclusion}\label{sec:Conclusion}
This paper presented a novel FL protocol, called Hybrid-FL, which extends FedCS to mitigate the non-IID data problem that degrades the model performance.
Hybrid-FL constructs an approximately IID dataset on the server by gathering the data from a limited number of clients who allow their data to be uploaded to the server, and the model updated using the IID data is aggregated with other models that are updated by other clients.
Furthermore, we designed heuristic algorithms to select the optimal set of model-uploading clients and the sets of clients and data to be uploaded.
Subsequently, we simulated an MEC environment and conducted experiments by performing realistic ML tasks to demonstrate the effective performance of our protocol.
Our experimental results revealed that Hybrid-FL with 1\% of data-uploading clients significantly improved the classification accuracy when the data was non-IID.
An interesting direction for future work is to consider the energy consumption of clients.

\section*{Acknowledgment}
This work was supported in part by JST ACT-I Grant Number JPMJPR17UK and KDDI Foundation.

\bibliographystyle{IEEEtran.bst}
\bibliography{./style/ICC.bib}

\begin{thebibliography}{10}
\providecommand{\url}[1]{#1}
\csname url@samestyle\endcsname
\providecommand{\newblock}{\relax}
\providecommand{\bibinfo}[2]{#2}
\providecommand{\BIBentrySTDinterwordspacing}{\spaceskip=0pt\relax}
\providecommand{\BIBentryALTinterwordstretchfactor}{4}
\providecommand{\BIBentryALTinterwordspacing}{\spaceskip=\fontdimen2\font plus
\BIBentryALTinterwordstretchfactor\fontdimen3\font minus
  \fontdimen4\font\relax}
\providecommand{\BIBforeignlanguage}[2]{{%
\expandafter\ifx\csname l@#1\endcsname\relax
\typeout{** WARNING: IEEEtran.bst: No hyphenation pattern has been}%
\typeout{** loaded for the language `#1'. Using the pattern for}%
\typeout{** the default language instead.}%
\else
\language=\csname l@#1\endcsname
\fi
#2}}
\providecommand{\BIBdecl}{\relax}
\BIBdecl

\bibitem{DBLP:conf/aistats/McMahanMRHA17}
B.~McMahan, E.~Moore, D.~Ramage, S.~Hampson, and B.~A. y~Arcas,
  ``Communication-efficient learning of deep networks from decentralized
  data,'' in \emph{Proc. AISTATS 2017}, Fort Lauderdale, FL, USA, Apr. 2017,
  pp. 1273--1282.

\bibitem{Nishio13}
T.~Nishio, R.~Shinkuma, T.~Takahashi, and N.~B. Mandayam, ``Service-oriented
  heterogeneous resource sharing for optimizing service latency in mobile
  cloud,'' in \emph{Proc. First International Workshop on Mobile Cloud
  Computing \& Networking}, Bangalore, India, 2013, pp. 19--26.

\bibitem{Zeng16}
D.~{Zeng}, L.~{Gu}, S.~{Guo}, Z.~{Cheng}, and S.~{Yu}, ``Joint optimization of
  task scheduling and image placement in fog computing supported
  software-defined embedded system,'' \emph{IEEE Trans. Computers}, vol.~65,
  no.~12, pp. 3702--3712, Dec. 2016.

\bibitem{10.1007/978-3-7908-2604-3_16}
L.~Bottou, ``Large-scale machine learning with stochastic gradient descent,''
  in \emph{Proc. COMPSTAT 2010}, Y.~Lechevallier and G.~Saporta, Eds.,
  Heidelberg, Germany, Aug. 2010, pp. 177--186.

\bibitem{nishio2018client}
T.~Nishio and R.~Yonetani, ``Client selection for federated learning with
  heterogeneous resources in mobile edge,'' in \emph{Proc. IEEE ICC}, May 2019,
  pp. 1--7.

\bibitem{zhao2018federated}
Y.~Zhao, M.~Li, L.~Lai, N.~Suda, D.~Civin, and V.~Chandra, ``Federated learning
  with non-{IID} data,'' \emph{CoRR}, vol. abs/1806.00582, Jun. 2018.

\bibitem{sattler2019robust}
F.~Sattler, S.~Wiedemann, K.-R. M{\"u}ller, and W.~Samek, ``Robust and
  communication-efficient federated learning from non-iid data,'' \emph{CoRR},
  vol. abs/1903.02891, Mar. 2019.

\bibitem{wang2019adaptive}
S.~Wang, T.~Tuor, T.~Salonidis, K.~K. Leung, C.~Makaya, T.~He, and K.~Chan,
  ``Adaptive federated learning in resource constrained edge computing
  systems,'' \emph{IEEE J. Sel. Areas Commun.}, vol.~37, no.~6, pp. 1205--1221,
  Jun. 2019.

\bibitem{yang2019scheduling}
H.~H. Yang, Z.~Liu, T.~Q. Quek, and H.~V. Poor, ``Scheduling policies for
  federated learning in wireless networks,'' \emph{CoRR}, vol. abs/1908.06287,
  Oct. 2019.

\bibitem{sesia2011lte}
S.~Sesia, M.~Baker, and I.~Toufik, \emph{LTE-the UMTS long term evolution: from
  theory to practice}.\hskip 1em plus 0.5em minus 0.4em\relax John Wiley \&
  Sons, 2011.

\bibitem{ITU-R}
M.~Series, ``Guidelines for evaluation of radio interface technologies for
  {IMT-Advanced},'' Report ITU-R M.2135-1, 2009.

\bibitem{6834753}
M.~R. Akdeniz, Y.~Liu, M.~K. Samimi, S.~Sun, S.~Rangan, T.~S. Rappaport, and
  E.~Erkip, ``Millimeter wave channel modeling and cellular capacity
  evaluation,'' \emph{IEEE J. Sel. Areas Commun.}, vol.~32, no.~6, pp.
  1164--1179, Jun. 2014.

\bibitem{krizhevsky2009learning}
A.~Krizhevsky and G.~Hinton, ``Learning multiple layers of features from tiny
  images,'' Technical report, Univ. of Toronto, 2009.

\bibitem{Konecny2016}
J.~Konecn{\'{y}}, H.~B. McMahan, F.~X. Yu, P.~Richt{\'{a}}rik, A.~T. Suresh,
  and D.~Bacon, ``Federated learning: Strategies for improving communication
  efficiency,'' in \emph{Proc. NIPS Workshop on Private Multi-Party Machine
  Learning}, Dec. 2016.

\bibitem{Xiao2017FashionMNISTAN}
H.~Xiao, K.~Rasul, and R.~Vollgraf, ``Fashion-mnist: a novel image dataset for
  benchmarking machine learning algorithms,'' \emph{CoRR}, vol. abs/1708.07747,
  Sep. 2017.

\end{thebibliography}

\end{document}